\documentclass[11pt]{article}

\usepackage[final]{acl}

\usepackage{times}
\usepackage{latexsym}

\usepackage[T1]{fontenc}

\usepackage[utf8]{inputenc}

\usepackage{microtype}

\usepackage{inconsolata}

\usepackage{graphicx}

\usepackage{subcaption}

\usepackage{amsmath,amssymb,bbm}

\usepackage[normalem]{ulem}
\useunder{\uline}{\ul}{}
\usepackage{booktabs} 

\usepackage{enumitem}

\title{Sigmoid Head for Quality Estimation under Language Ambiguity}

\author{Tu Anh Dinh \and Jan Niehues \\
        Karlsruhe Institute of Technology \\ Karlsruhe, Germany \\ \texttt{tu.dinh@kit.edu, jan.niehues@kit.edu}}

\begin{document}
\maketitle
\begin{abstract}
Language model (LM) probability is not a reliable quality estimator, as natural language is ambiguous. When multiple output options are valid, the model's probability distribution is spread across them, which can misleadingly indicate low output quality. This issue is caused by two reasons: (1) LMs' final output activation is softmax, which does not allow multiple correct options to receive high probabilities simultaneuously and (2) LMs' training data is single, one-hot encoded references, indicating that there is only one correct option at each output step. We propose training a module for Quality Estimation on top of pre-trained LMs to address these limitations. The module, called Sigmoid Head, is an extra unembedding head with sigmoid activation to tackle the first limitation. To tackle the second limitation, during the negative sampling process to train the Sigmoid Head, we use a heuristic to avoid selecting potentially alternative correct tokens. Our Sigmoid Head is computationally efficient during training and inference. The probability from Sigmoid Head is notably better quality signal compared to the original softmax head. As the Sigmoid Head does not rely on human-annotated quality data, it is more robust to out-of-domain settings compared to supervised QE.
\end{abstract}

\section{Introduction}
\textbf{Quality Estimation} (QE) is the task of providing a score estimation of the model output quality during inference when the ground-truth output is not available. 
The most straightforward way is to use model probability: it is computationally lightweight, does not require any additional modules, as the model probability comes for free during generation.
However, previous works have shown that, for language generation tasks, model probability is not a good signal of output quality, largely due to the inherent ambiguity of natural language \cite{ott2018analyzing,stahlberg-kumar-2022-jam,fadeeva-etal-2024-fact,flores-etal-2025-improving,dinh-niehues-2025-generative}. 
For a given input, multiple outputs can be valid, making the model probability spread out more even when the output is high quality. For convenience, we call this the \textbf{\textit{ambiguity-induced underconfidence}} issue.

\textit{Ambiguity-induced underconfidence} is partly caused by the architecture and training setup of current language models (LMs). First, the LMs' final output activation is softmax, which enforces the probabilities of all output options to sum to one. Second, the training target is a single reference that is one-hot encoded, which teaches the model that only one output option is correct. As a result, multiple valid options cannot all receive high probabilities at the same time.

We propose to address the above issue as follows: we train an additional unembedding layer (a.k.a. \textit{"projection layer"} or \textit{"LM head"}) on top of pre-trained LMs to perform Quality Estimation. The output activation of this layer is sigmoid instead of softmax, so each output option is modeled independently. 
We use the same training data used for the original LM to train our module. To address the disadvantage of the single-reference, one-hot encoding setup, we propose a heuristic for negative sampling during training to avoid selecting potentially alternative correct tokens as negative. We refer to the module as the \textbf{Sigmoid Head}.

In short, our contributions are as follows:
\begin{itemize}[nolistsep,leftmargin=*]
    \item We identify LMs' architecture and training issues causing \textbf{\textit{ambiguity-induced underconfidence}}.
    \item We propose a Sigmoid Head \footnote{Implementation available at \url{https://github.com/TuAnh23/sigmoid-head-qe}.} on top of pre-trained LMs for Quality Estimation to address these issues. The module is trained on the same data as standard generative training, requires no additional labeled data, and is computationally lightweight during training and inference.
    \item Our experiments show that the probabilities from the Sigmoid Head provide better quality signals than those from the standard softmax head. Moreover, since our method does not rely on human-labeled quality data for training, it is more robust and outperforms the supervised COMET Kiwi \cite{rei-etal-2022-cometkiwi} on domain-specific machine translation (biomedical). It also outperforms common QE approaches: Monte Carlo Sequence Entropy \cite{malinin2020uncertainty,kuhn2023semantic} and LLM Self Judge.
\end{itemize}

\section{Background and Motivation}
\subsection{Background: Standard LM Training}
\paragraph{Training Objective}
Text-generation language models (LMs) are trained auto-regressively. Given an input sequence $\mathbf{x}=(x_1,\ldots,x_{|\mathbf{x}|})$ and a previously generated output prefix $\mathbf{y}_{<i}=(y_1,\ldots,y_{i-1})$, the model with parameters $\theta$ is trained to predict the next token at time step $i$:
\begin{equation*}
P_{\theta}(y_i \mid \mathbf{x}, \mathbf{y}_{<i}) \xrightarrow{\text{train}} \hat{P}(y_i \mid \mathbf{x}, \mathbf{y}_{<i}),
\end{equation*}
where the target distribution $\hat{P}$ is a one-hot vector that assigns probability 1 to the single reference token and 0 to all others. We exclude label smoothing from our explanation for simplicity (See Appendix \ref{app:ce_loss_d2} for more details).

\paragraph{Model Architecture}
Most text-generation models consist of a transformer (encoder-decoder or decoder-only)
with parameters $\theta_{\text{tr}}$, which produces a hidden representation
$\mathbf{h}_i \in \mathbb{R}^d$ at generation step~$i$. This (last) hidden state is mapped to
vocabulary-sized logits $\mathbf{z}^{\text{out}}_i $ via an unembedding head with parameters $\theta_{\text{out}}$:
\begin{equation*}
\mathbf{z}^{\text{out}}_i = W_{\text{out}} \mathbf{h}_i + \mathbf{b}_{\text{out}},
\end{equation*}
where $W_{\text{out}} \in \mathbb{R}^{|\mathcal{V}| \times d}$ and
$\mathbf{b}_{\text{out}} \in \mathbb{R}^{|\mathcal{V}|}$ ($\mathbf{b}_{\text{out}}$ is optional).
A softmax function is then applied to obtain a distribution over the vocabulary
$\mathcal{V}$:
\begin{equation*}
P_{\theta}(y_i \mid \mathbf{x}, \mathbf{y}_{<i})
= \mathrm{softmax}(\mathbf{z}^{\text{out}}_i),
\end{equation*}
where $\theta = \{\theta_{\text{tr}}, \theta_{\text{out}}\}$.

\subsection{Motivation: Issues of Standard LMs} \label{sec:motivation}

Natural language is ambiguous, where there are often multiple ways to express the same meaning. As a result, there are usually multiple correct output options at each generation step. The standard training above does not faciliate this in two aspects:
\begin{itemize}[nolistsep,leftmargin=*]
    \item \textbf{D1}: The softmax function enforces mutual exclusivity, i.e., probabilities must sum to one. When multiple tokens are correct, they cannot all receive high probability.
    \item \textbf{D2}: Training relies on a single one-hot reference at each step, teaching the model that exactly one token is correct and all others are incorrect.
    \label{disadvantages}
\end{itemize}

While \textbf{D1} could be addressed by replacing the final softmax function with sigmoid, \textbf{D2} remains problematic. To further demonstrate \textbf{D2}, consider the following example where there are 2 samples in a translation training dataset (without loss of generality to other text-generation tasks such as instruction following or language modeling):

\begin{quote}
\textbf{Input}: Ich werde anfangen \\
\textbf{Output 1}: I will \textit{begin} \\
\textbf{Output 2}: I will \textit{start}
\end{quote}

For the same input and output prefix, the model receives conflicting one-hot encoded target supervision. Sample 1 provides the target:
\begin{align*}
\hat{P}(y_i=\text{``begin''} \mid \mathbf{x}, \mathbf{y}_{<i}) &= 1, \\
\hat{P}(y_i\neq\text{``begin''} \mid \mathbf{x}, \mathbf{y}_{<i}) &= 0,
\end{align*}
while Sample 2 provides:
\begin{align*}
\hat{P}(y_i=\text{``start''} \mid \mathbf{x}, \mathbf{y}_{<i}) &= 1, \\
\hat{P}(y_i\neq\text{``start''} \mid \mathbf{x}, \mathbf{y}_{<i}) &= 0.
\end{align*}

Sample 1 encourages the model to place all probability mass on “begin” and none on “start”, while Sample 2 does the opposite. If the model assign probability close to 1 to ``begin'', the loss will be low for Sample 1 but very high for Sample 2, where “begin” is an incorrect target. Similar argument holds for ``start''. To reduce the total loss, the model cannot assign a probability close to 1 to either token. Instead, it must distribute probability mass between them. This applied to both models with softmax final activation and models with sigmoid final activation. See Appendix \ref{app:ce_loss_d2} for detailed behavior of the cross entropy loss in this example.

\begin{figure*}
    \centering
    \includegraphics[width=0.85\linewidth]{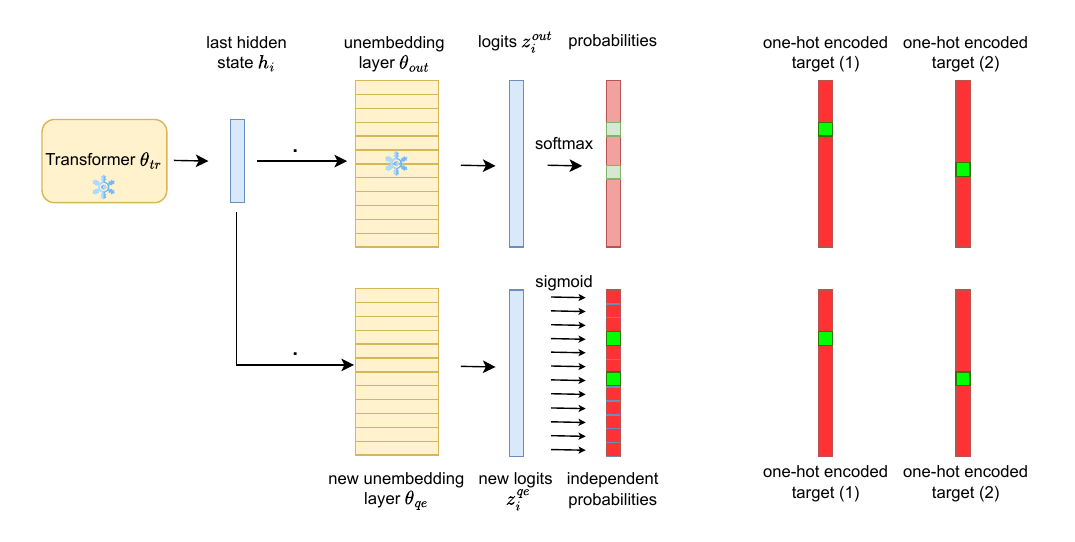}
    \caption{Extended language model (LM) architecture with our proposed Sigmoid Head. The weights of the original components from the LM are kept unchanged. We initialize the Sigmoid Head from the original softmax head, and train it to predict independent scores for each token in the vocabulary.}
    \label{fig:architecture}
\end{figure*}

\paragraph{Implicit Effect of Ambiguity} Language ambiguity is learned implicitly by language models. As discussed above, when the training data shows that multiple outputs are correct, the model learns to spread probability mass across these tokens. This behavior was also observed by \citet{dinh-niehues-2025-generative}. They show that, specifically for ambiguous generation tasks, multiple tokens in the softmax distribution often receive relatively high probability compared to the remaining tokens at each output step. They refer to these tokens as "dominant tokens", which are likely to correspond to valid alternative outputs. They propose a heuristic to identify such tokens. We adopt this idea in our method, described in Section \ref{sec:method}.

\subsection{Noise-Contrastive Estimation}
An alternative to the final softmax activation in standard LM training is to use the sigmoid function, as proposed by \citet{Mnih2012AFA}, known as Noise-Contrastive Estimation (NCE). NCE reformulates next-token training as a binary discrimination task: the reference token is treated as a positive example, while several tokens sampled from a predefined noise distribution are treated as negatives. The model is trained to assign high probability to the reference token and low probability to the sampled noise tokens using a sigmoid-based objective. The goal is to avoid the need for full softmax calculation during training and reduces computation for large vocabularies.

As the softmax function may not be well suited to model the ambiguity of language (Section \ref{sec:motivation}), we make use of NCE in our method, described in Section \ref{sec:method}.

\section{Proposed: Sigmoid Head for QE} \label{sec:method}

We propose a lightweight module for Quality Estimation on top of a trained generative model that explicitly accounts for ambiguity in the training data. The weights $\theta = \{\theta_{\text{tr}}, \theta_{\text{out}}\}$ of the generative model are kept unchanged. We introduce an additional
unembedding head with parameters $\theta_{\text{qe}}$, operating on the same hidden
states $\mathbf{h}_i$ produced by the transformer $\theta_{\text{tr}}$, and trained on the same single-reference, one-hot encoded target as the standard setup of the original model (Figure \ref{fig:architecture}).

\paragraph{Sigmoid Instead of Softmax}
To address the disadvantage \textbf{\hyperref[disadvantages]{D1}}, for our module, instead of using softmax as the output activation, we use the sigmoid function, similar to Noise-Contrastive Estimation (NCE) \cite{Mnih2012AFA}. We then refer to our module as the \textbf{Sigmoid Head}. Concretely, given the transformer hidden
state $\mathbf{h}_i$, the new unembedding head $\theta_{\text{qe}}$ computes the logits
\begin{equation*}
\mathbf{z}^{\text{qe}}_i = W_{\text{qe}} \mathbf{h}_i + \mathbf{b}_{\text{qe}},
\end{equation*}
where $W_{\text{qe}} \in \mathbb{R}^{|\mathcal{V}| \times d}$ and
$\mathbf{b}_{\text{qe}} \in \mathbb{R}^{|\mathcal{V}|}$ (optional) are trainable parameters.
An element-wise sigmoid function is then applied:
\begin{equation*}
P_{\theta'}(y_i \mid \mathbf{x}, \mathbf{y}_{<i}) = \sigma(\mathbf{z}^{\text{qe}}_i)
\end{equation*}
where $\theta' = \{\theta_{\text{tr}}, \theta_{\text{qe}}\}$.

Unlike softmax, sigmoid removes the constraint that probabilities must sum to one. Multiple tokens can simultaneously receive high probability, better matching settings where multiple outputs are correct.

At each output step $i$, we train the Sigmoid Head to distinguish correct from incorrect tokens, given the one-hot encoded target. Let $y_i^\ast$ denote the single reference token in the one-hot encoded target. The reference token is treated as a positive example with label 1. Negative examples are sampled from the non-reference tokens in the vocabulary $\mathcal{V} \setminus \{y_i^\ast\}$. We sample 10 negative tokens, as prior work has shown that this is an effective choice for different tasks \cite{NIPS2013_9aa42b31,mikolov-etal-2018-advances}. Negative tokens are sampled from a configurable distribution. Unless stated otherwise, we use token-frequency-based sampling, where tokens that appear more often in the training data are more likely to be selected as negatives. In this way, a token is expected to be selected as a negative as often as it appears as a positive. This balances the training signal and reduces bias toward frequent tokens. In Section~\ref{sec:neg_sampling_effect}, we evaluate alternative negative sampling strategies and analyze their impact on performance.

\paragraph{Ambiguity-Informed Negative Sampling: Avoid Dominant Tokens}
Recall disadvantage \textbf{\hyperref[disadvantages]{D2}}, where some non-reference tokens from $\mathcal{V} \setminus \{y_i^\ast\}$ may be valid output alternatives. To reduce the risk of sampling such tokens as negatives, we make use of the observation from \citet{dinh-niehues-2025-generative}: in a trained LM, the tokens with dominant probability mass in the softmax distribution are likely to be the alternative correct options. We therefore apply their heuristic (See Appendix \ref{app:find_dominant} for details) to \textbf{identify the set of dominant tokens} $\mathcal{D}_i$ from the original softmax distribution $P_{\theta}(y_i \mid \mathbf{x}, \mathbf{y}_{<i}) = \mathrm{softmax}(\mathbf{z}^{\text{out}}_i)$ and \textbf{exclude them from negative sampling}. Our approach can be viewed as a form of implicit knowledge distillation, as we are using the knowledge from the pretrained transformer $\theta_{\text{tr}}$ and softmax head $\theta_{\text{out}}$ to train our new sigmoid head $\theta_{\text{qe}}$. Formally, during training of the sigmoid head, negative samples $\mathcal{N}_i$ are drawn from
\begin{equation*}
\mathcal{V} \setminus \left( \{y_i^\ast\} \cup \mathcal{D}_i \right),
\end{equation*}
Intuitively, for potentially correct tokens in $\mathcal{D}_i$, the learning signal is postponed: the model neither learn that they are correct nor incorrect at this step.

The Sigmoid Head is trained with binary cross-entropy loss over the selected samples, consisting of the reference token $y_i^\ast$ and the negative
samples $\mathcal{N}_i$. The loss at output step $i$ is:
\begin{equation*}
\mathcal{L}_i = 
-\log p_i(y_i^\ast)
- \sum_{v \in \mathcal{N}_i} \log \bigl(1 - p_i(v)\bigr),
\end{equation*}

where $p_i(v)$ is the probability that the Sigmoid Head assigned to token $v \in \mathcal{V}$ at time step $i$.

During inference, the output text is generated by the original LM components $\theta = \{\theta_{\text{tr}}, \theta_{\text{out}}\}$ without modification. The Sigmoid Head $\theta_{\text{qe}}$ operates in parallel by taking the last hidden states from $\theta_{\text{tr}}$ at each generation step and produces a quality score for each output token.

Our Sigmoid Head has several advantages:
\begin{itemize}[nolistsep,leftmargin=*]
    \item We use the same training data as standard generative modeling and require no extra annotations. Since we do not rely on human-annotated quality scores, unlike supervised QE, our approach is expected to be more robust across domains.
    \item Training is computationally efficient thanks to negative sampling: at each generation step, the loss is computed over the reference token and a small set of sampled negative tokens, rather than over the full vocabulary. Consequently, only the corresponding rows of the unembedding matrix are updated. This efficiency is similar to NCE, although in our case it is not the main motivation.
    \item Inference is computationally efficient, as the computation of the final hidden state is shared between the softmax head and the Sigmoid Head. 
\end{itemize}

\section{Experimental Setup} 

\subsection{Models and Training Data}
Details of the language models (LMs) used in our paper and their corresponding training data can be found in Table \ref{tab:models}. We train Transformer Base from scratch using the base configuration of Fairseq \cite{ott2019fairseq} on 5M samples of ParaCrawl English-German Machine Translation data, filtered by Bicleaner AI \cite{zaragoza-bernabeu-etal-2022-bicleaner,de-gibert-etal-2024-new}. DeltaLM is finetuned on the same ParaCrawl data. Tower Instruct v2 and OLMo SFT v2 models are taken off the shelves. We experiment with adding our Sigmoid Head for Quality Estimation onto these models. More details on our motivation for model choices are in Appendix \ref{app:model_choices}. We discuss the increase in model parameter counts after adding the Sigmoid Head in Appendix \ref{app:params_count}.

\begin{table}[!htbp]
\small
\centering
\begin{tabular}{@{}lll@{}}
\toprule
Model                                        & Nr. Params                               & Train Data*                                     \\ \midrule
Transformer Base                             & \multicolumn{1}{l}{0.062B}                  & ParaCrawl en-de 5M                                    \\
\cite{ott2019fairseq}                        &                                          & \cite{banon-etal-2020-paracrawl}                \\
DeltaLM                                      & \multicolumn{1}{l}{0.830B}                & ParaCrawl en-de 5M                                    \\
\cite{ma2021deltalm}                         &                                          & \cite{banon-etal-2020-paracrawl}                \\
Tower Instruct v2                            & \multicolumn{1}{l}{7B}                   & TowerBlocks v2                                  \\
\cite{tower_llm_2024}                        &                                          & \cite{tower_llm_2024}                           \\
OLMo SFT v2                                  & \multicolumn{1}{l}{1B}                   & Tulu 3 SFT                                      \\
\cite{olmo20242olmo2furious}                 &                                          & \cite{lambert2024tulu3}                         \\ \midrule
\multicolumn{3}{l}{\begin{tabular}[c]{@{}l@{}}*: Training data from the last training phase of the model.\end{tabular}}
\end{tabular}
\caption{Models used in our paper, along with their training data from the last training phase. We use the same data to train our Sigmoid Heads.}
\label{tab:models}
\end{table}

\subsection{Test Data and Evaluation} \label{sec:exp_tasks}

We evaluate the Quality Estimation performance of our Sigmoid Head on three different generation tasks: Machine Translation, Paraphrase, and Question Answering. The QE score is obtained for each model output by multiplying the token-level scores produced by the Sigmoid Head.
We place more focus on the Machine Translation (MT) task, as MT evaluation is more well-studied, with different established data, models, and benchmarks thanks to the WMT Shared Tasks \cite{kocmi-etal-2024-findings,freitag-etal-2024-llms,zerva-etal-2024-findings}.

\paragraph{Machine Translation} For MT, we use ParaCrawl and WMT22 test sets for the two self-trained models as they are on the sentence level. We use WMT24 and the domain-specific BioMQM test sets \cite{zouhar-etal-2024-fine} for the other models. We evaluate two QE settings: using the LM to produce a quality score for its own translation, which we denote as "Eval Self", and using the LM to force-decode other translations and produce quality scores for the other translation, which we denote as "Eval Other". Human quality annotations are used as the ground-truth when available (Eval Other). Otherwise, for Eval Self, we use pseudo ground-truth produced by the reference-based XCOMET \cite{guerreiro-etal-2024-xcomet}, as \citet{dinh-etal-2024-quality} has shown that reference-based methods are robust enough to rank reference-free QE. We report Pearson correlation, which measures how the QE scores linearly correlate with the ground truth quality.

\paragraph{Other Tasks} For paraphrasing, we use the PAWS-X dataset \cite{yang-etal-2019-paws} and evaluate both Eval Self and Eval Other settings. For question answering, we use GSM8k (Math domain) \cite{cobbe2021training} and TruthfulQA (generic domain) \cite{lin-etal-2022-truthfulqa} and only evaluate Eval Self. For Eval Self, pseudo ground-truth is generated by prompting Qwen2.5 72B Instruct \cite{qwen2.5} to evaluate whether the model output is correct or not. In these tasks, ground-truth or pseudo ground-truth labels are binary, thus we report on Binary Cross Entropy (BCE) loss, which measures how close the QE scores are to the ground-truth binary labels, without the need to define a threshold.

More details on the test sets and choice of language pairs can be found in Appendix \ref{app:testsets}.

\paragraph{Baselines} We compare our approach to the standard softmax probability and BoostedProb \cite{dinh-niehues-2025-generative}. BoostedProb is an adjusted version of the softmax probability, which partially addresses \textit{ambiguity-induced underconfidence}. 
Specifically, the tokens with dominant probability mass in the softmax distribution are assumed to be the likely correct output alternatives. 
The confidence of these tokens is then boosted by having the total probability mass of the whole dominant set as the quality score.

For MT, we use the supervised COMET Kiwi \cite{rei-etal-2022-cometkiwi} as an upper baseline, as it is trained on human-labeled quality data as opposed to our approach. We also compare our approach to common unsupervised QE approaches. The first one is Monte Carlo sequence entropy \cite{malinin2020uncertainty,kuhn2023semantic}, where we sample 10 output sequences for each input and compute sequence-level probability entropy. The second one is LLM Self Judge, in which we prompt the model to evaluate its own output.

\section{Results and Discussion}

\subsection{Overall Performance} \label{sec:main_exp}
\paragraph{QE performance on self-generated translation output (Eval Self)} Table~\ref{tab:all_self_gen_mt} shows the QE results on self-generated outputs from the Paracrawl, WMT22, and WMT24 translation datasets across different models and language pairs. Overall, our Sigmoid Head consistently outperforms the standard softmax head and the BoostedProb baseline. In many cases, it still underperforms the supervised COMET Kiwi. However, COMET Kiwi results might be biased, as we use XCOMET outputs as the gold quality scores, which belong to the same model family as COMET Kiwi.

\begin{table}[!htbp]
\small
\centering
\setlength{\tabcolsep}{4pt}
\begin{tabular}{@{}lcccc@{}}
\toprule
                               & Softmax & Boosted        & Sigmoid              & COMET                \\
                               & Head    & Prob           & Head                 & Kiwi                 \\ \midrule
{\ul \textbf{TransformerBase}} &         &                &                      &                      \\
ParaCrawl                      & 0.155   & 0.235          & \textbf{0.383}       & {\ul \textbf{0.536}} \\
WMT22 en-de                    & 0.199   & 0.367          & \textbf{0.586}       & {\ul \textbf{0.722}} \\
{\ul \textbf{DeltaLM}}         &         &                & \textbf{}            & {\ul \textbf{}}      \\
ParaCrawl                      & 0.131   & 0.218          & \textbf{0.403}       & {\ul \textbf{0.537}} \\
WMT22 en-de                    & 0.165   & 0.291          & \textbf{0.515}       & {\ul \textbf{0.634}} \\
{\ul \textbf{Tower}}           &         &                & \textbf{}            & {\ul \textbf{}}      \\
WMT 24 en-de                   & 0.148   & 0.414          & \textbf{0.468}       & {\ul \textbf{0.562}} \\
WMT 24 en-fr                   & 0.155   & 0.370          & \textbf{0.391}       & {\ul \textbf{0.530}} \\
WMT 24 en-es                   & 0.183   & \textbf{0.446} & 0.415                & {\ul \textbf{0.525}} \\
WMT 24 en-pt                   & 0.150   & 0.458          & \textbf{0.472}       & {\ul \textbf{0.533}} \\
WMT 24 en-nl                   & 0.145   & 0.419          & \textbf{0.513}       & {\ul \textbf{0.552}} \\
WMT 24 en-it                   & 0.154   & 0.394          & \textbf{0.448}       & {\ul \textbf{0.588}} \\
WMT 24 en-ko                   & 0.165   & 0.595          & \textbf{0.566}       & {\ul \textbf{0.626}} \\
WMT 24 en-cn                   & 0.160   & 0.491          & {\ul \textbf{0.537}} & \textbf{0.500}       \\
WMT 24 en-ru                   & 0.176   & 0.505          & \textbf{0.524}       & {\ul \textbf{0.625}} \\
{\ul \textbf{Olmo}}            &         &                & \textbf{}            &                      \\
WMT 24 en-de                   & 0.195   & 0.398          & {\ul \textbf{0.606}} & \textbf{0.588}       \\
WMT 24 en-es                   & 0.209   & 0.508          & {\ul \textbf{0.672}} & \textbf{0.613}       \\ \bottomrule
\multicolumn{5}{l}{\begin{tabular}[c]{@{}l@{}}Best scores are shown as \textbf{\underline{<score>}}, second best are \textbf{<score>}.\end{tabular}}
\end{tabular}
\caption{QE performance on models' self-generated translations (Eval Self) in Pearson Correlation $\uparrow$.}
\label{tab:all_self_gen_mt}
\end{table}

\paragraph{QE performance on others' translation output (Eval Others)} The QE results on scoring others' translations are shown in Table~\ref{tab:all_fd_mt}. As before, the Sigmoid Head outperforms the standard softmax head and the BoostedProb baseline, but still lags behind the supervised COMET Kiwi model on the WMT test sets. However, COMET Kiwi is trained on in-domain WMT shared task data, which gives it an advantage over our unsupervised approach. This is confirmed by the results on BioMQM, a domain-specific test set, where our method generally outperforms COMET Kiwi, indicating better robustness on out-of-domain settings.

Our approach also works on providing QE scores on the word level, allowing DeltaLM to outperform supervised QE. Details are in Appendix \ref{app:word_level}.

\paragraph{QE performance on paraphrasing and question answering} Table \ref{tab:other_tasks} shows the QE performance on tasks other than translations: paraphrasing and question answering. Our Sigmoid Head consistently outperforms Softmax Head and BoostedProb across different languages and test sets. 

We provide a qualitative analysis in Appendix \ref{app:qualitative}, presenting example cases where the original Softmax Head fails but the Sigmoid Head succeeds.

\paragraph{Comparison to common QE approaches}
Table~\ref{tab:fancy_baselines} compares the QE performance of our Sigmoid Head with Monte Carlo Sequence Entropy and LLM Self Judge. Monte Carlo Sequence Entropy is usually computed from log probabilities produced by the standard softmax head. We also report a variant that uses log probabilities from our Sigmoid Head. Overall, our Sigmoid Head consistently outperforms both the standard Monte Carlo Sequence Entropy baseline and the LLM Self Judge baseline. The Monte Carlo Sequence Entropy variant based on the Sigmoid Head either slightly degrades performance or gives negligible improvement over using the Sigmoid Head alone, while being more expensive because it requires multiple sampled sequences. LLM Self Judge performs poorly in most cases. This is because Tower is a specialized model that is not trained for general LLM-as-a-Judge tasks beyond machine translation evaluation, and OLMo lacks strong LLM-as-a-Judge capability due to its small size. The only exception is Tower Self Judge on the WMT24 translation data, since the model is trained for this task. These results highlight a limitation of LLM Self Judge: it does not work well in general unless the model is sufficiently large and not overly specialized for a small set of tasks.

\begin{table}[!htbp]
\centering
\small
\setlength{\tabcolsep}{3pt}
\begin{tabular}{@{}lcccc@{}}
\toprule
                               & Softmax & Boosted              & Sigmoid              & COMET                \\
                               & Head    & Prob                 & Head                 & Kiwi                 \\ \midrule
{\ul \textbf{TransformerBase}} &         &                      &                      &                      \\
WMT 22 en-de                   & 0.076   & 0.131                & \textbf{0.166}       & {\ul \textbf{0.365}} \\
                               &         &                      &                      &                      \\
{\ul \textbf{DeltaLM}}         &         &                      &                      &                      \\
WMT 22 en-de                   & 0.081   & 0.141                & \textbf{0.201}       & {\ul \textbf{0.365}} \\
                               &         &                      &                      &                      \\
{\ul \textbf{Tower}}           &         &                      &                      &                      \\
WMT 24 en-de                   & 0.094   & 0.215                & \textbf{0.272}       & {\ul \textbf{0.349}} \\
WMT 24 en-es                   & 0.056   & 0.170                & \textbf{0.230}       & {\ul \textbf{0.429}} \\
WMT 24 en-zh                   & 0.011   & 0.114                & \textbf{0.161}       & {\ul \textbf{0.385}} \\
WMT 24 en-ru                   & 0.072   & 0.180                & \textbf{0.205}       & {\ul \textbf{0.345}} \\
{\ul \textbf{Average}}         & 0.058   & 0.170                & \textbf{0.217}       & {\ul \textbf{0.377}} \\
                               &         &                      &                      &                      \\
BioMQM en-pt                   & 0.067   & \textbf{0.214}       & {\ul \textbf{0.263}} & 0.078                \\
BioMQM en-es                   & 0.150   & \textbf{0.300}       & {\ul \textbf{0.314}} & 0.150                \\
BioMQM en-fr                   & 0.145   & \textbf{0.364}       & {\ul \textbf{0.369}} & 0.238                \\
BioMQM en-de                   & 0.114   & {\ul \textbf{0.450}} & \textbf{0.441}       & 0.194                \\
BioMQM en-it                   & 0.093   & {\ul \textbf{0.343}} & \textbf{0.268}       & 0.235                \\
BioMQM en-zh                   & 0.055   & 0.223                & {\ul \textbf{0.316}} & \textbf{0.274}       \\
BioMQM en-ru                   & 0.065   & {\ul \textbf{0.301}} & \textbf{0.300}       & 0.207                \\
{\ul \textbf{Average}}         & 0.098   & \textbf{0.314}       & {\ul \textbf{0.325}} & 0.197                \\ \bottomrule
\multicolumn{5}{l}{\begin{tabular}[c]{@{}l@{}}Best scores are shown as \textbf{\underline{<score>}}, second best are \textbf{<score>}.\end{tabular}}
\end{tabular}
\caption{QE performance when forced-decoding on other translations (Eval Others) in Pearson $\uparrow$.}
\label{tab:all_fd_mt}
\end{table}

\begin{table}[!htbp]
\small
\centering
\setlength{\tabcolsep}{4pt}
\begin{tabular}{@{}lllrcc@{}}
\toprule
Mode        & Model & Lang. & \multicolumn{1}{c}{Softmax} & Boosted & Sigmoid        \\
            &       &       & \multicolumn{1}{c}{Head}    & Prob    & Head           \\ \midrule
\multicolumn{6}{l}{{\ul \textit{Paraphrasing (Pawsx)}}}                              \\
Eval Self   & Tower & en    & 11.668                      & 1.925   & \textbf{1.056} \\
            &       & de    & 9.025                       & 1.645   & \textbf{0.422} \\
            &       & es    & 9.334                       & 1.362   & \textbf{0.399} \\
            &       & fr    & 8.847                       & 1.256   & \textbf{0.325} \\
            &       & zh    & 12.745                      & 2.925   & \textbf{0.622} \\
            & Olmo  & en    & 14.684                      & 3.769   & \textbf{0.883} \\
Eval Others & Tower & en    & 10.835                      & 6.858   & \textbf{4.546} \\
            &       & de    & 11.756                      & 8.726   & \textbf{3.151} \\
            &       & es    & 10.868                      & 7.324   & \textbf{2.309} \\
            &       & fr    & 9.977                       & 6.297   & \textbf{1.849} \\
            &       & zh    & 12.362                      & 9.257   & \textbf{2.172} \\
            & Olmo  & en    & 13.364                      & 9.096   & \textbf{2.832} \\
\multicolumn{6}{l}{{\ul \textit{Question Answering Math (GSM8K)}}}                   \\
Eval Self   & Olmo  & en    & 12.802                      & 0.843   & \textbf{0.642} \\
\multicolumn{6}{l}{{\ul \textit{Question Answering Generic (TruthfulQA)}}}           \\
Eval Self   & Olmo  & en    & 9.424                       & 6.219   & \textbf{1.698} \\ \bottomrule
\end{tabular}
\caption{QE performance on tasks other than translation, meansured in Binary Cross Entropy loss (BCE) $\downarrow$.}
\label{tab:other_tasks}
\end{table}

\begin{table}[!h]
\small
\centering
\setlength{\tabcolsep}{3pt}
\begin{tabular}{@{}llrrrr@{}}
\toprule
Test Data  & \multicolumn{1}{c}{Model} & \multicolumn{1}{c}{Sigmoid} & \multicolumn{2}{c}{Monte Carlo}                                       & \multicolumn{1}{c}{LLM}   \\
           & \multicolumn{1}{c}{}      & \multicolumn{1}{c}{Head}    & \multicolumn{2}{c}{Seq. Entropy}                                      & \multicolumn{1}{c}{Self}  \\
           & \multicolumn{1}{c}{}      & \multicolumn{1}{c}{}        & \multicolumn{1}{c}{{\ul Softmax}} & \multicolumn{1}{c}{{\ul Sigmoid}} & \multicolumn{1}{c}{Judge} \\ \midrule
\multicolumn{6}{l}{{\ul \textit{QE performance in Pearson $\uparrow$}}}                                                                                                  \\
WMT24      & Tower                     & \textbf{0.468}              & 0.093                             & 0.441                             & 0.293                     \\
           & Olmo                      & \textbf{0.606}              & 0.106                             & 0.552                             & 0.093                     \\
           &                           &                             &                                   &                                   &                           \\
\multicolumn{6}{l}{{\ul \textit{QE performance in BCE $\downarrow$}}}                                                                                                    \\
PAWS-X     & Tower                     & 1.056                       & 11.828                            & \textbf{1.047}                    & 22.993                    \\
           & Olmo                      & \textbf{0.883}              & 16.228                            & 0.920                             & 6.416                     \\
GSM8k      & Olmo                      & \textbf{0.642}              & 15.241                            & 0.798                             & 16.408                    \\
TruthfulQA & Olmo                      & 1.698                       & 9.681                             & \textbf{1.674}                    & 24.729                    \\ \bottomrule
\end{tabular}
\caption{Comparison to common QE approaches.}
\label{tab:fancy_baselines}
\end{table}

\subsection{Effect of Negative Sampling Strategies} \label{sec:neg_sampling_effect}

\begin{figure}[!h]
    \centering
    \begin{subfigure}{0.48\linewidth}
        \centering
        \includegraphics[width=\linewidth]{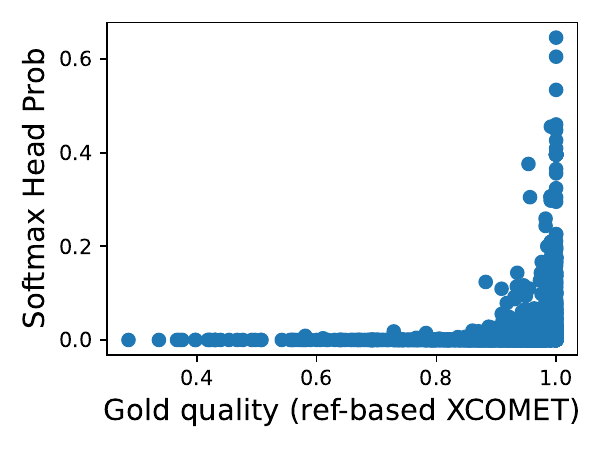}
        \caption{Original Softmax Head \\ \hspace*{0.45cm}(Baseline)}
        \label{fig:raw_prob}
    \end{subfigure} \hfill
    \begin{subfigure}{0.48\linewidth}
        \centering
        \includegraphics[width=\linewidth]{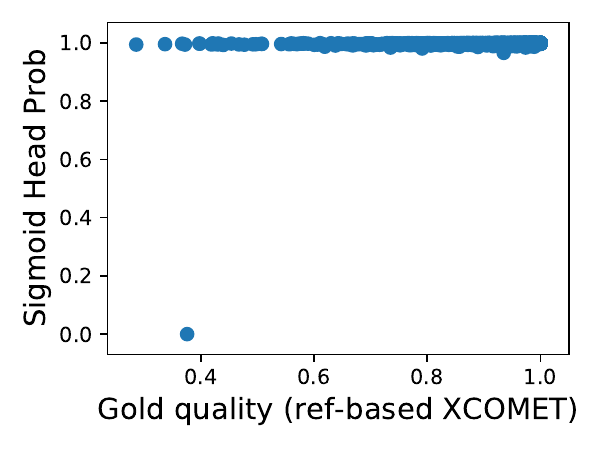}
        \caption{Sigmoid Head\\ \hspace*{0.45cm}NS: Random}
        \label{fig:sigmoid_random}
    \end{subfigure}

    \begin{subfigure}{0.48\linewidth}
        \centering
        \includegraphics[width=\linewidth]{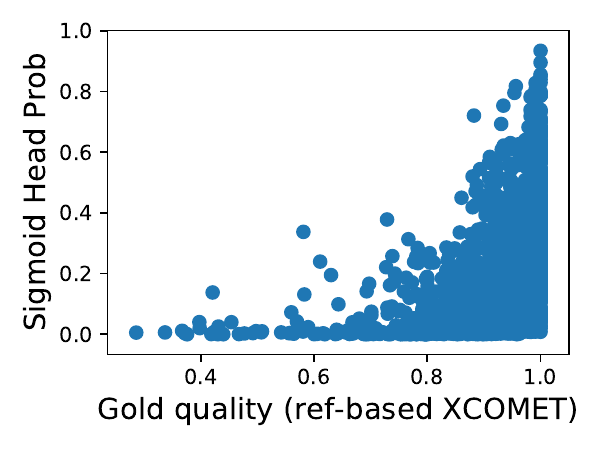}
        \caption{Sigmoid Head\\ \hspace*{0.45cm}NS: From Softmax Head\vspace{0.37cm}}
        \label{fig:sigmoid_softmax}
    \end{subfigure} \hfill
    \begin{subfigure}{0.48\linewidth}
        \centering
        \includegraphics[width=\linewidth]{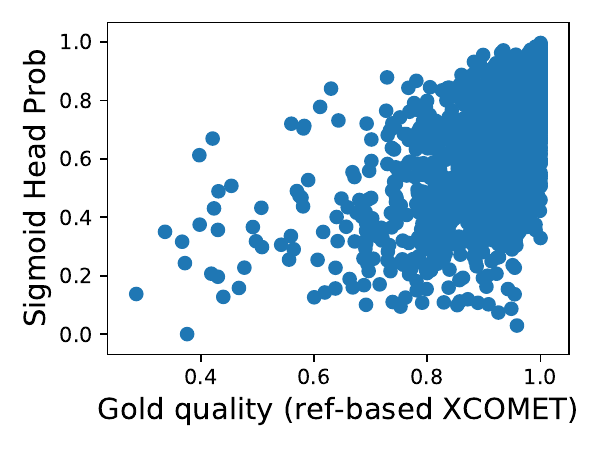}
        \caption{Sigmoid Head\\ \hspace*{0.45cm}NS: From Softmax Head,\\ \hspace*{0.95cm} avoid dominant}
        \label{fig:sigmoid_softmax_avoid}
    \end{subfigure}

        \begin{subfigure}{0.48\linewidth}
        \centering
        \includegraphics[width=\linewidth]{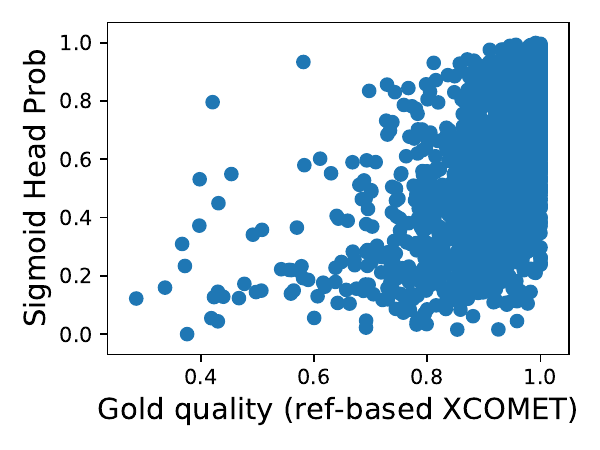}
        \caption{Sigmoid Head\\ \hspace*{0.45cm}NS: Token Frequency\vspace{0.37cm}}
        \label{fig:sigmoid_token_freq}
    \end{subfigure} \hfill
    \begin{subfigure}{0.48\linewidth}
        \centering
        \includegraphics[width=\linewidth]{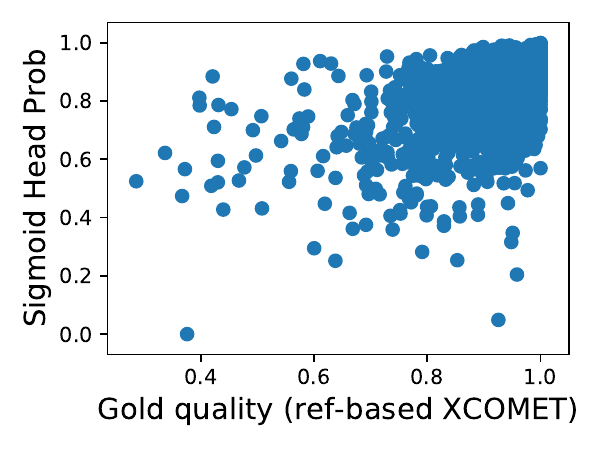}
        \caption{Sigmoid Head\\ \hspace*{0.45cm}NS: Token Frequency,\\ \hspace*{0.95cm} avoid dominant}
        \label{fig:sigmoid_token_freq_avoid}
    \end{subfigure}

    \caption{Ground-truth versus predicted quality scores.}
    \label{fig:self_gen_fairseq}
\end{figure}

We perform an ablation study to see how different negative sampling strategies affect the QE performance of the Sigmoid Head. 
We run these experiments on Transformer-base and DeltaLM models, as these models allow multiple training runs with reasonable computational cost. Figure~\ref{fig:self_gen_fairseq} plots ground-truth quality scores against predicted quality scores for different model settings.

\paragraph{Standard Softmax: Underconfidence}
Figure~\ref{fig:raw_prob} shows the behavior of the standard softmax head. We observe the \textit{ambiguity-induced underconfidence} issue: outputs with high gold quality (points on the right side of the plot) have predicted probabilities spread across the full range from 0 to 1.

\begin{table*}[!htbp]
\centering
\small
\setlength{\tabcolsep}{3pt}
\begin{tabular}{llcccc}
\toprule 
                           &                                          & \multicolumn{2}{c}{Transformer Base}        & \multicolumn{2}{c}{DeltaLM}                 \\
                           &                                          & ParaCrawl            & WMT22                & ParaCrawl            & WMT22                \\ \midrule
Probability (Softmax Head) &                                          & 0.155                & 0.199                & 0.131                & 0.165                \\
BoostedProb                &                                          & 0.235                & 0.367                & 0.218                & 0.291                \\
                           & {\ul \textit{Negative sampling}}         &                      &                      &                      &                      \\
Sigmoid Head               & Random                                   & 0.123                & 0.163                & 0.141                & 0.209                \\
                           & Token Freq                               & 0.304                & 0.521                & 0.142                & 0.209                \\
                           & Token Freq + Avoid Dominant              & 0.380                & 0.588                & 0.394                & 0.508                \\
                           & Softmax                                  & 0.197                & 0.441                & 0.126                & 0.357                \\
                           & Softmax + Avoid Dominant                 & 0.326                & 0.580                & 0.239                & 0.467                \\
                           & Softmax t2 + Avoid Dominant              & 0.373                & \textbf{0.602}       & 0.387                & 0.510                \\
                           & Softmax t2 + Token Freq + Avoid Dominant & \textbf{0.383}       & 0.586                & \textbf{0.403}       & \textbf{0.515}       \\
COMET Kiwi                 &                                          & {\ul \textbf{0.536}} & {\ul \textbf{0.722}} & {\ul \textbf{0.537}} & {\ul \textbf{0.634}} \\
\bottomrule
\multicolumn{6}{l}{\begin{tabular}[c]{@{}l@{}}Best scores are shown as \textbf{\underline{<score>}}, second best are \textbf{<score>}.\end{tabular}}
\end{tabular}
\caption{Effect of different negative sampling strategies on QE performance.}
\label{tab:self_gen_fairseq}
\end{table*}

\paragraph{Random Negative Sampling}
Figure~\ref{fig:sigmoid_random} shows the Sigmoid Head trained with random negative sampling. The probability is overconfident: outputs of all quality levels receive probabilities close to 1. This happens for two main reasons. First, at each step, incorrect tokens greatly outnumber correct ones. Random sampling therefore mostly picks clearly wrong tokens, which gives a weak learning signal and does not teach the model to distinguish correct tokens from plausible but incorrect ones. Second, random sampling treats all tokens as negatives equally often, while positive examples depend on token frequency in the training data. As a result, frequent tokens appear more often as positives than as negatives, biasing the model to assign them high scores even when they are incorrect.

\paragraph{Negative Sampling from the Softmax Head}
Figure~\ref{fig:sigmoid_softmax} shows the Sigmoid Head trained with negative sampling from the softmax head distribution, where highly ranked tokens (except the reference) are more likely to be sampled as negatives. In this case, the sigmoid head remains underconfident, similar to the baseline in Figure~\ref{fig:raw_prob}, although the effect is weaker. This is because alternatively correct tokens are still likely to be sampled as negatives, which exposes the model to the same \textit{ambiguity-induced underconfidence} issue.

\paragraph{Ambiguity-Informed Negative Sampling: Avoid Dominant Tokens}
The \textit{ambiguity-induced underconfidence} issue is mitigated when we explicitly exclude dominant tokens from negative sampling. As shown in Figure~\ref{fig:sigmoid_softmax_avoid}, the predicted probabilities better align with gold quality.

\paragraph{Negative Sampling from Token Frequency}
Similar trends are observed when negative sampling is based on token frequency in the training data. Figure~\ref{fig:sigmoid_token_freq} shows that frequency-based sampling alone still suffers from underconfidence. When dominant tokens are explicitly excluded (Figure~\ref{fig:sigmoid_token_freq_avoid}), the issue is mitigated, and the confidence estimates become more aligned with gold quality.

\paragraph{Quantitative Results}
The above observations are confirmed by the QE performance results in terms of Pearson correlation, shown in Table~\ref{tab:self_gen_fairseq}. The proposed Sigmoid Head consistently outperforms the standard softmax head and the BoostedProb baseline. The best-performing settings use negative sampling based on token frequency or the softmax distribution, combined with explicit exclusion of dominant tokens. We further experiment with applying a temperature of 2 to the softmax distribution used for negative sampling. This reduces the impact of \textit{ambiguity-induced underconfidence} and leads to the best overall results. This best setting is used for Transformer Base and DeltaLM in the main experiments (Section \ref{sec:main_exp}),

For Olmo and Tower, we only evaluate a small number of top settings identified from the Transformer-based and DeltaLM experiments, due to their higher computational cost for training. Our proxy runs show that negative sampling from token frequency alone, with dominant tokens excluded, performs best for these models. We therefore use this setting in the main experiment (Section \ref{sec:main_exp}) for OLMo and Tower.


\section{Related Work}
In the field of Machine Translation (MT), the Quality Estimation task is well studied, with the well-established method of training a separate module on human-labeled quality scores on MT output, with the representative example of COMET-kiwi \cite{rei-etal-2022-cometkiwi}. However, outside of MT, there is a lack of such training data to build such supervised QE modules. This calls for the need of unsupervised QE approaches, which are well aligned with \textit{Uncertainty Quantification} techniques. These techniques often fall into several categories \cite{fadeeva-etal-2023-lm}: information-based approaches which make use of the model probability \cite{fomicheva-etal-2020-unsupervised}, 
ensemble-based approaches which require generating multiple outputs \cite{kuhn2023semantic,dinh-niehues-2023-perturbation}, self-validation approaches which prompt the LLMs to evaluate themselves \cite{kadavath2022language}, and density-based approaches which require access to the model training data to detect out-of-distribution instances during inference \cite{NEURIPS2018_abdeb6f5,ren2023outofdistribution}. 
Amongst these categories, information-based approaches are the most straightforward and computationally lightweight, as model probabilities come for free at generation.

Previous works have pointed out that model probability for text generation tasks is not a good signal of output quality due to the inherent ambiguity of natural language, which we called the \textbf{ambiguity-induced underconfidence} issue \cite{ott2018analyzing,stahlberg-kumar-2022-jam,fadeeva-etal-2024-fact,flores-etal-2025-improving,dinh-niehues-2025-generative}. 
\citet{fadeeva-etal-2024-fact} address this issue by detecting tokens that are synonyms and contradicting compared to the final selected token, and re-calculate the quality score given the probabilities of only those tokens.
\citet{flores-etal-2025-improving} address the issue by, instead of only looking at the probability of the output as a quality signal, they calculate the probability ratio between the highest-ranked sequences and the rest, or evaluate the thinness of the distribution’s tail. 
\citet{dinh-niehues-2025-generative} addresses the issue by assigning dominant tokens in the softmax distribution a quality score equal to the total probability mass of all dominant tokens, rather than their individual probability mass. 
These works address \textit{ambiguity-induced underconfidence} by modifying the softmax probability distributions.

In this paper, we address \textit{ambiguity-induced underconfidence} at the training architectural level: instead of softmax, we use the sigmoid activation, so that each output option is modeled independently. In this regard, noise contrastive estimation (NCE) \cite{gutmann2010noise,Mnih2012AFA} and SCONES \cite{stahlberg-kumar-2022-jam} are the closest to our work, since they also switch out the softmax activation for sigmoid. However, NCE's focus is to train models on a large vocabulary size more efficiently with negative sampling so that not the whole unembedding layer needs to be learnt at every update step; SCONES' focus is to improve generation quality. Unlike us, both works do not take into account the \textit{ambiguity-induced underconfidence} issue, i.e., all non-reference options can still be sampled as negative, even when they could be the alternative correct options.


\section{Conclusion}
This work addressed the \textit{ambiguity-induced underconfidence} issue of LM probability for QE by introducing the Sigmoid Head, an additional unembedding head with sigmoid activation trained on top of frozen pretrained models. The sigmoid activation allows multiple output options to receive high confidence, while ambiguity-informed negative sampling avoids penalizing likely alternative correct tokens during training. The Sigmoid Head is trained on the same data as standard language modeling, requires no human-annotated quality labels, and is computationally efficient at both training and inference time. Across machine translation, paraphrasing, and question answering, the probability from Sigmoid Head provides a stronger quality signal than the standard softmax head and common unsupervised QE baselines. As it does not rely on supervised quality data, it is also more robust in out-of-domain settings and can outperform supervised QE models under domain shift.

\section*{Limitations}
The main motivation of this work is to improve the training setup and model architecture to better account for the ambiguity of natural language. In standard large language model training, such ambiguity already exists during next-token prediction pretraining, where multiple tokens can be valid at each generation step. However, due to limited computational resources, we only apply our proposed training setup to the final instruction-following training stage. Applying our approach during the pretraining phase could potentially bring stronger benefits, as the model could learn ambiguity more effectively from the large-scale pretraining data. Another limitation is that we evaluate our approach on only two LLMs, Tower Instruct 7B and OLMo 1B, as these are among the few model families for which the training data is publicly available. Lastly, adding the Sigmoid Head increases the parameter count of language models, though the resulting overhead is relatively minor for most state-of-the-art large models.

\section*{Acknowledgements}
This work was supported by the Helmholtz Programme-oriented Funding, with project number 46.24.01, project name AI for Language Technologies. 
We acknowledge the HoreKa supercomputer funded by the Ministry of Science, Research and the Arts Baden-Wurttemberg and by the Federal Ministry of Education and Research. This work also received support from the European Union’s Horizon research and innovation programme under grant agreement No 101135798, project Meetween (My Personal AI Mediator for Virtual MEETtings BetWEEN People).

\bibliography{custom}

\appendix

\section{Cross-entropy Loss Behaviour on Multi-target Samples}
\label{app:ce_loss_d2}

Recall the example in Section \ref{sec:motivation} where there are 2 samples in the training data:

\begin{quote}
\textbf{Input}: Ich fange an \\
\textbf{Output 1}: I will \textit{begin} \\
\textbf{Output 2}: I will \textit{start}
\end{quote}

which lead to the two training targets:
\begin{align*}
\hat{P}(y_i=\text{``begin''} \mid \mathbf{x}, \mathbf{y}_{<i}) &= 1, \\
\hat{P}(y_i\neq\text{``begin''} \mid \mathbf{x}, \mathbf{y}_{<i}) &= 0,
\end{align*}
and
\begin{align*}
\hat{P}(y_i=\text{``start''} \mid \mathbf{x}, \mathbf{y}_{<i}) &= 1, \\
\hat{P}(y_i\neq\text{``start''} \mid \mathbf{x}, \mathbf{y}_{<i}) &= 0.
\end{align*}

When the output activation function is the standard softmax, the loss commonly used in practice is referred to as the negative log-likelihood (NLL), which is equivalent to the categorical cross-entropy loss with one-hot target:
\begin{equation*}
\mathcal{L}_{\text{CE}} = - \sum_v^{\mathcal{V}} t_v \log p_v,
\end{equation*}
where $t_v$ is the one-hot target and $p_v$ is the predicted probability. When $p_{\text{begin}} \approx 1$, $p_{\text{start}}$ is enforced to be close to 0 given the softmax. The loss value is then low for Sample 1. However, for Sample 2 where $t_{\text{start}} = 1$, the loss value will be very large:

\begin{align*}
\mathcal{L}_{\text{CE}} &= - \sum_v^{\mathcal{V}} t_v \log p_v \\
&= - \sum_{v \neq \text{start}}^{\mathcal{V}} t_v \log p_v - t_{\text{start}} \log p_{\text{start}}  \\
&\approx - \sum_{v \neq \text{start}}^{\mathcal{V}} t_v \log p_v - 1 \times \log 0 \\
&\approx \infty.
\end{align*}
Similarly for the case when the model assigns $p_{\text{start}} \approx 1$. In practice, label smoothing is often applied to alleviate this issue to some extent by redistributing a small amount of probability mass from the ground-truth token to other tokens in the target distribution. However, label smoothing does not fundamentally resolve the problem: it still enforces a single preferred target per sample and penalizes assigning high probability to alternative valid outputs, only less severely. As a result, the model is still discouraged from assigning high probability to multiple correct tokens simultaneously.

This issue persists even when we switch out the softmax output function with sigmoid. Consider the binary cross-entropy loss commonly used with the sigmoid output function:
\begin{equation*}
\mathcal{L}_{\text{BCE}} = - \sum_v^{\mathcal{V}} \left[ t_v \log p_v + (1 - t_v)\log(1 - p_v) \right].
\end{equation*}

Again, if $p_{\text{begin}} \approx 1$, the loss is low for Sample 1 but very high for Sample 2 where $t_{\text{begin}} = 0$:
\begin{align*}
&\mathcal{L}_{\text{BCE}} = - \sum_v^{\mathcal{V}} \left[ t_v \log p_v + (1 - t_v)\log(1 - p_v) \right] \\
&= - \sum_{v \neq \text{begin}}^{\mathcal{V}} \left[ t_v \log p_v + (1 - t_v)\log(1 - p_v) \right] \\
& \hspace{0.43cm}- t_{\text{begin}} \log p_{\text{begin}} - (1 - t_{\text{begin}}) \log(1 - p_{\text{begin}}) \\
&= - \sum_{v \neq \text{begin}}^{\mathcal{V}} \left[ t_v \log p_v + (1 - t_v)\log(1 - p_v) \right] \\
& \hspace{0.43cm}- 1 \times \log 0 \\
&\approx \infty .
\end{align*}

Similarly for the case when the model assigns $p_{\text{start}} \approx 1$. Therefore, even with sigmoid, one-hot target supervision teaches the model to distribute probability mass amongst valid alternatives, preventing it from assigning high confidence to multiple correct tokens.

\section{Finding Dominant Tokens} \label{app:find_dominant}
\citet{dinh-niehues-2025-generative} show that, at an output step, tokens with dominant probability mass in the softmax distribution (so-called ``dominant tokens'') tend to be correct alternative outputs. They propose a heuristic to identify these tokens. First, the probability distribution is sorted in descending order. Then, moving from high to low probabilities, they search for a ``significant drop.'' A drop is considered significant if it exceeds both a relative threshold ($x=30\%$ of the preceding probability) and an absolute threshold $\epsilon=0.005$. This drop separates the dominant tokens from the remaining ones. In our work, we apply this heuristic to identify dominant tokens and exclude them from negative sampling, since they are potentially correct alternatives.

\section{Motivation for Model Choices} \label{app:model_choices}

We selected four models to add the Sigmoid Head to: Transformer Base (trained from scratch), DeltaLM (fine-tuned), Tower, and OLMo (off-the-shelf). As mentioned in Section~\ref{sec:exp_tasks}, we focus on Machine Translation (MT), since MT evaluation is better studied than evaluation for other tasks. We therefore trained the encoder-decoder Transformer Base and DeltaLM on a medium-sized MT dataset (ParaCrawl 5M) to enable faster experiments with different negative sampling methods. We chose Tower because it is a state-of-the-art decoder-only LLM specialized for MT. We chose OLMo, a general-purpose language model, to test whether the Sigmoid Head also works for generation tasks beyond MT.

An important factor is that we want to train our Sigmoid Head on the exact same data that was used to train the original language model. This is another reason for choosing Tower and OLMo, as their training data is publicly available. We use the instruction-tuned versions of these two models to reduce the computational cost of training the Sigmoid Head, since instruction-tuning data is typically much smaller than language modeling data.

\section{Increase in Parameter Count with Sigmoid Head} \label{app:params_count}
We report parameter counts before and after adding the Sigmoid Head in Table \ref{tab:model_sizes}. Introducing an additional unembedding head increases memory usage. This overhead is more pronounced for models with relatively small total parameter counts but large vocabularies (e.g., DeltaLM and OLMo 1B). However, for many state-of-the-art large models, the unembedding layer constitutes only a small fraction of the total parameters, for example, 5.87\% for OLMo-3-7B, 2.77\% for gpt-oss-20B, 0.496\% for gpt-oss-120B, and 0.265\% for Qwen3-235B-A22B. In such cases, the additional memory cost of the Sigmoid Head is relatively minor.

\begin{table}[h!]
\centering
\small
\setlength{\tabcolsep}{3pt}
\begin{tabular}{lccc}
\hline
\textbf{Model} & \textbf{Base Model} & \textbf{Base +} & \textbf{Increase} \\
& \textbf{Size} & \textbf{Sigmoid Head} & \textbf{(\%)} \\
\hline
Transformer Base & 0.062B & 0.068B & 9.7\% \\
DeltaLM & 0.830B & 1.086B & 30.8\% \\
Tower Instruct v2 & 7.000B & 7.131B & 1.9\% \\
OLMo SFT v2 & 1.000B & 1.206B & 20.6\% \\
\hline
\end{tabular}
\caption{Model parameter count before and after adding the Sigmoid Head.}
\label{tab:model_sizes}
\end{table}

\section{Test Sets for Evaluation} \label{app:testsets}
Table \ref{tab:testset_stats} shows the statistics of the test sets used in our experiments. More details of the test sets are as follows.

\paragraph{WMT and BioMQM} The WMT and BioMQM machine translation test sets contain source sentences and reference translations for multiple language pairs. They also include system output translations from WMT shared task participants, together with human quality scores. We use these scores as ground truth for quality estimation, which enables evaluation in the \textit{Eval Other} mode.

\paragraph{PAWS-X} Each sample in PAWS-X is a pair of sentences $(S_1, S_2)$ along with binary labels: 1 if the two sentences are paraphrases, 0 otherwise. This also enables our assessment of the "Eval Other" mode, where we treat $S_1$ as the input text, $S_2$ as the candidate output text, and the binary labels as the ground truth quality score. In contrast, for the "Eval Self" mode, we let the LM generate its own paraphrase of $S_1$, and generate pseudo ground truth quality score with Qwen2.5 72B Instruct.

\paragraph{GSM8k and TruthfulQA} Each GSM8k example contains a math problem and its ground-truth answer. Each TruthfulQA example contains a question, a set of correct answers, and a set of incorrect answers. For these two datasets, we only evaluate the \textit{Eval Self} mode. Pseudo ground-truth quality scores are generated using Qwen2.5 72B Instruct. To improve the reliability of the pseudo ground-truth, we provide all available reference information to Qwen2.5 72B Instruct: the ground-truth answer for GSM8k, and the correct and incorrect answer lists for TruthfulQA.

\paragraph{Language selection} For the multilingual translation test sets (WMT24 and BioMQM), for Tower, we include all the available English-X language pairs that Tower was trained on: German, French, Spanish, Dutch, Italian, Korean, Chinease and Russian. For OLMo, since it is claimed to be English-centric and not meant for translation, we only include two high-resource language pairs: English-German and English-Spanish. Similarly, for the multilingual PAWS-X paraphrase test set, we evaluate all languages supported by Tower. For OLMo, we evaluate only English.

\begin{table}[!htbp]
\small
\setlength{\tabcolsep}{4pt}
\centering
\begin{tabular}{@{}lllcc@{}}
\toprule
Task                                            & Test Set                                      & Lang.                                    & Nr. In $^1$                                    & Nr. Out $^2$                                    \\ \midrule
MT                                              & ParaCrawl                                     & en-de                                    & 5000                                           & -                                               \\
                                                & WMT22                                         & en-de                                    & 2037                                           & 10980                                           \\
                                                & WMT24                                         & en-de                                    & 960                                            & 8262                                            \\
                                                &                                               & en-fr                                    & 960                                            & -                                               \\
                                                &                                               & en-es                                    & 960                                            & 8242                                            \\
                                                &                                               & en-pt                                    & 960                                            & -                                               \\
                                                &                                               & en-nl                                    & 960                                            & -                                               \\
                                                &                                               & en-it                                    & 960                                            & -                                               \\
                                                &                                               & en-ko                                    & 960                                            & -                                               \\
                                                &                                               & en-cn                                    & 960                                            & 7608                                            \\
                                                &                                               & en-ru                                    & 960                                            & 8242                                            \\
                                                & BioMQM                                        & en-pt                                    & 935                                            & 449                                             \\
                                                &                                               & en-es                                    & 717                                            & 2188                                            \\
                                                &                                               & en-fr                                    & 614                                            & 2456                                            \\
                                                &                                               & en-de                                    & 757                                            & 2710                                            \\
                                                &                                               & en-it                                    & 1040                                           & 514                                             \\
                                                &                                               & en-zh                                    & 639                                            & 4437                                            \\
                                                &                                               & en-ru                                    & 550                                            & 1650                                            \\
Paraphrasing                                    & PAWS-X                                        & en                                       & 1749                                           & 1749                                            \\
                                                &                                               & de                                       & 1972                                           & 1972                                            \\
                                                &                                               & es                                       & 1976                                           & 1976                                            \\
                                                &                                               & fr                                       & 1958                                           & 1958                                            \\
                                                &                                               & zh                                       & 1962                                           & 1962                                            \\
QA                                              & GSM8k                                         & en                                       & 1319                                           & -                                               \\
                                                & TruthfulQA                                    & en                                       & 817                                            & -                                               \\ \bottomrule
\multicolumn{5}{l}{\begin{tabular}[c]{@{}l@{}}$^1$: Number of input samples, which is the test size for \\Eval Self.\\ $^2$: Number of candidate outputs, which is the test size for \\Eval Others.\end{tabular}}
\end{tabular}
\caption{Statistics of test sets.}
\label{tab:testset_stats}
\end{table}

\section{Word-level QE} \label{app:word_level}

We evaluate our Sigmoid Head on providing quality scores on the word level. We use the HJQE test set \cite{yang-etal-2023-rethinking}, which contains source sentences and participants' translations from the WMT20 QE Shared Task \cite{specia-etal-2020-findings-wmt} along with human-annotated quality labels (OK/BAD) on the word level. We again do force-decoding to score these translations (Eval Others). As the labels are binary, we report Binary Cross Entropy loss (BCE). As an upper baseline, we use the supervised WMT 21 OpenKiwi model \cite{specia2021findings,kim-etal-2017-predictor}.

The results are shown in Table \ref{tab:word_level}. Our Sigmoid Head again outperforms the Softmax Head and BoostedProb baselines. Sigmoid Head on top of the small Transformer Base model underperforms compared to the supervised QE. However, Sigmoid Head on top of DeltaLM enables DeltaLM to slightly outperform supervised QE, which was not the case for Softmax Head and BoostedProb.

\begin{table}[htbp]
\centering
\small
\setlength{\tabcolsep}{2pt}
\begin{tabular}{@{}lccclc@{}}
\toprule
Model            & Softmax & Boosted & Sigmoid        &  & Supervised     \\
                 & Head    & Prob    & Head           &  & QE             \\ \midrule
Transformer Base & 1.580   & 0.984   & 0.513          &  & \textbf{0.332} \\
DeltaLM          & 1.362   & 0.829   & \textbf{0.329} &  & 0.332          \\ \bottomrule
\end{tabular}
\caption{QE performance on word-level, measured in BCE loss $\downarrow$ (Eval Others).}
\label{tab:word_level}
\end{table}

\section{Qualitative Analysis: When Does Softmax Fail?} \label{app:qualitative}

\begin{table*}[t!]
\centering
\setlength{\tabcolsep}{2pt}
\small
\begin{tabular}{p{3cm} p{3cm} p{3cm} c c c p{2.8cm}}
\hline
\textbf{Input sentence} & \textbf{Gold translation} & \textbf{MT output} & \textbf{Human} & \textbf{Sigmoid} & \textbf{Softmax} & \textbf{Softmax scores of} \\
& & & \textbf{score} & \textbf{score} & \textbf{score} & \textbf{valid alternatives} \\
\hline

Kurze Notiz über die Presse – der Freitag &
Short note about the press – Friday &
\textbf{Brief} note on the press – Friday &
0.98 & 0.99 & 0.02 &
Short 0.40, A 0.34, \textbf{Brief} 0.19, Quick 0.04 \\ \\

Transgender: Identifiziert sich nicht mit dem bei der Geburt zugewiesenen Geschlecht. &
Transgender: Does not identify with the gender assigned at birth. &
Transgender: Does not identify with the \textbf{sex} assigned at birth. &
0.99 & 0.98 & 0.02 &
gender 0.74, \textbf{sex} 0.20 \\ \\

Haushaltspolitik: Steuereinnahmen niedriger als erwartet &
Budget policy: Tax revenues lower than expected &
\textbf{Budgetary} policy: Tax revenue lower than expected &
0.92 & 0.98 & 0.00 &
Budget 0.52, \textbf{Budgetary} 0.01 \\ \\

Funktion: Weiterschalten auf nächsten Programmschritt (Anzeige) &
Function: Advance to next programming step (display) &
Function: \textbf{Switch} to the next program step (display) &
0.97 & 0.99 & 0.00 &
Proceed 0.12, Continue 0.10, Move 0.10, Go 0.09, Advance 0.08, \textbf{Switch} 0.05 \\ \\

Schnelle Lieferung und gute Verarbeitung &
Fast delivery and good manufacturing &
Fast delivery and good \textbf{workmanship} &
0.96 & 0.98 & 0.19 &
processing 0.57, \textbf{workmanship} 0.30, handling 0.03, crafting 0.02, manufacturing 0.02 \\ \\

Ekelhafter, dauerhafter Geruch &
Disgusting, permanent smell &
Disgusting, \textbf{persistent} \textbf{odor} &
1.00 & 0.99 & 0.07 &
\textbf{persistent} 0.35, permanent 0.25, lasting 0.12; smell 0.51, \textbf{odor} 0.03, odour 0.01 \\

\hline
\end{tabular}
\caption{Example of good MT outputs along with their QE scores from human, Sigmoid Head, and Softmax Head.}
\label{tab:qualitative}
\end{table*}

We provide example sentences in Table \ref{tab:qualitative} where the Softmax Head fails while the Sigmoid Head succeeds. Through manual inspection, we identify many cases where the translations are adequate, but the Softmax Head assigns very low scores because one or more tokens have multiple valid alternatives. We highlight such tokens in bold. In the last column, we report the softmax probabilities of the top valid alternatives (word-level scores), showing that these tokens compete for probability mass under softmax, leading to ambiguity-induced underconfidence. The data used is WMT23 German–English, and the scoring model is Tower Instruct v2.

\section{Hardware}
Training the Sigmoid Head on top of LLMs (Tower and OLMo) is conducted on a single H100 GPU with 96\,GB memory. All other processes are run on a single A100 GPU with 40\,GB memory, including training the Transformer Base and DeltaLM models and performing inference for all models.

\section{License For Artifacts}
The license for artifacts used in our paper is as follows:

\begin{itemize}
    \item ParaCrawl dataset \cite{banon-etal-2020-paracrawl}: Creative Commons CC0
    \item WMT22 dataset \cite{kocmi-etal-2022-findings}: Apache License 2.0
    \item WMT24 dataset \cite{kocmi-etal-2024-findings}: Apache License 2.0
    \item BioMQM \cite{zouhar-etal-2024-fine}: Apache License 2.0
    \item GSM8k dataset \cite{cobbe2021training}: MIT License
    \item TruthfulQA dataset \cite{lin-etal-2022-truthfulqa}: Apache License 2.0 
    \item DeltaLM model \cite{ma2021deltalm}: MIT License
    \item Tower model \cite{tower_llm_2024}: CC BY NC 4.0
    \item OLMo model \cite{olmo20242olmo2furious}: Apache License 2.0
\end{itemize}

\end{document}